\documentclass[conference]{IEEEtran}
\ifCLASSINFOpdf
\else
\fi
\usepackage{amsmath}
\usepackage{pifont}
\ifCLASSOPTIONcompsoc
  \usepackage[caption=false,font=normalsize,labelfont=sf,textfont=sf]{subfig}
\else
  \usepackage[caption=false,font=footnotesize]{subfig}
\fi
\usepackage{url}
\usepackage{hyperref}[]
\hypersetup{
    colorlinks=true,
    linkcolor=blue,
    filecolor=magenta,      
    urlcolor=cyan,
    pdftitle={Overleaf Example},
    pdfpagemode=FullScreen,
    hidelinks % this deletes the colors
    }

\usepackage{cleveref} 
\usepackage{tcolorbox}
\usepackage{geometry}
\usepackage{multicol}
\usepackage{listings}

\usepackage{blindtext}
\PassOptionsToPackage{dvipsnames}{xcolor} % extra color names https://www.overleaf.com/learn/latex/Using_colors_in_LaTeX#Accessing_additional_named_colors
\usepackage{xcolor}

\crefname{lstlisting}{listing}{listings}
\Crefname{lstlisting}{Listing}{Listings}

\lstdefinelanguage{Prolog}{
    morekeywords={:-, ?, @, <, >, =, is},
    keywordstyle=\color{blue}, % color for keywords
    basicstyle=\scriptsize\ttfamily, % smaller font for code
    morecomment=[l]\%, % line comment character
    commentstyle=\color{green!50!black}, % color for comments
    morestring=[b]', % single quoted strings
    stringstyle=\color{red}, % color for strings
}

\usepackage{tcolorbox}
\usepackage{tikz}
\usepackage{lipsum} % For dummy text, remove if unnecessary
\tcbuselibrary{listingsutf8} % Allows handling LaTeX code within box

\newtcolorbox{promptbox}[1][]{
    fontupper=\sffamily\footnotesize,
    colback=yellow!5!white,
    colframe=yellow!50!black,
    colbacktitle=yellow!75!black,    
    box align=top,
    #1 % This allows extra parameters to be passed when the box is called
}

\usepackage{float} % For float customization
\usepackage{newfloat} % For defining custom float environments

\DeclareFloatingEnvironment[fileext=lop]{Box}
\crefname{Box}{box}{boxes}
\Crefname{Box}{Box}{Boxes}

\usepackage{capt-of}
\usepackage{caption} 
\usepackage{soul}

\usepackage[nolist,nohyperlinks]{acronym}
\newacro{CNN}{Convolutional Neural Network}
\newacro{CNNs}{Convolutional Neural Networks}
\newacro{DNN}{Deep Neural Network}
\newacro{GPS}{Global Positioning System}
\newacro{GNSS}{Global Navigation Satellite System}
\newacro{NLOS}{non-line-of-sight}
\newacro{PC}{Point Cloud}
\newacro{ADAS}{Advanced Driver Assistance System}
\newacro{LIDAR}[LiDAR]{Light Detection And Ranging}
\newacro{HD map}{High Definition map}
\newacro{EV}{Embedding Vector}
\newacro{SLAM}{Simultaneouos Localization And Mapping}
\newacro{MLP}{MultiLayer Perceptron}
\newacro{IMU}{Inertial Measurement Unit}
\newacro{ML}{Machine Learning}
\newacro{SfM}{Structure from Motion}
\newacro{PnP}{Perspective-n-Points}
\newacro{ASPP}{Atrous Spatial Pyramid Pooling}
\newacro{SVM}{Support Vector Machine}
\newacro{CRF}{Conditional Random Field}
\newacro{RNN}{Recursive Neural Network}
\newacro{HUMANAV}{Human-like Semantic Navigation for Autonomous Vehicles}
\newacro{LLM}{Large Language Model}
\newacro{ASP}{Answer Set Programming}
\newacro{KR}{Knowledge Representation and Reasoning}
\newacro{KB}{Knowledge Base}
\newacro{RAG}{Retrieval Augmented Generation}

\usepackage{graphicx}
\usepackage{amssymb}

\IEEEoverridecommandlockouts 

\begin{document}
\title{Human-like Semantic Navigation for Autonomous Driving using Knowledge Representation and Large Language Models}

\author{Augusto Luis Ballardini~\IEEEmembership{Member,~IEEE} and Miguel Ángel Sotelo~\IEEEmembership{Fellow,~IEEE}
\thanks{
All authors are with the Computer Engineering Department, University of Alcalá, Alcalá de Henares, 
Madrid, Spain. Email: 
\href{mailto://augusto.ballardini@uah.es}{augusto.ballardini@uah.es},
}}

\markboth{Journal of \LaTeX\ Class Files,~Vol.~14, No.~8, August~2015}%
{Shell \MakeLowercase{\textit{et al.}}: Bare Demo of IEEEtran.cls for IEEE Journals}

\maketitle

\begin{abstract}
Achieving full automation in self-driving vehicles remains a challenge, especially in dynamic urban environments where navigation requires real-time adaptability. Existing systems struggle to handle navigation plans when faced with unpredictable changes in road layouts, spontaneous detours, or missing map data, due to their heavy reliance on predefined cartographic information.
In this work, we explore the use of Large Language Models to generate Answer Set Programming rules by translating informal navigation instructions into structured, logic-based reasoning. ASP provides non-monotonic reasoning, allowing autonomous vehicles to adapt to evolving scenarios without relying on predefined maps.
We present an experimental evaluation in which LLMs generate ASP constraints that encode real-world urban driving logic into a formal knowledge representation. By automating the translation of informal navigation instructions into logical rules, our method improves adaptability and explainability in autonomous navigation.
Results show that LLM-driven ASP rule generation supports semantic-based decision-making, offering an explainable framework for dynamic navigation planning that aligns closely with how humans communicate navigational intent.
\end{abstract}

% \begin{IEEEkeywords}
% Self-Driving, Localization, Knowledge-Representation, KR, Scene Understanding, Urban Settings, Semantics, Answer Set Programming, Explainability.)
% \end{IEEEkeywords}


% \fi
\IEEEpeerreviewmaketitle

\section{Introduction}
\noindent
This work explores the development of a semantic navigation system for autonomous vehicles, proposing and implementing an approach that moves beyond traditional metric-based localization.
Autonomous driving and assistance systems are set to transform the future of mobility. 
While fully autonomous vehicles are not yet a reality, \acp{ADAS} represent the most advanced commercially deployed solutions available today. 
%These systems rely on sensor-acquired data, such as cameras and \ac{LIDAR} sensors, and have rapidly evolved over the past decade, driven by technological advancements and the integration of machine learning techniques and neural networks.
These systems rely on sensor data, such as cameras and \ac{LIDAR}, and have evolved rapidly over the past decade, driven by advances in machine learning and neural network technologies.
\ac{ADAS} currently represent the state of the art in commercial systems, performing best on high-speed roads (highways and expressways), where vehicle interactions are relatively simple. 
On the other hand, urban environments require a much more advanced understanding of the surroundings, as the complexity of traffic interactions and routing dynamics requires a level of reliability that systems relying solely on sensor data or pre-trained algorithms often cannot guarantee.
\setlength{\belowcaptionskip}{-15pt}
\begin{figure}[t]
\centering
\includegraphics[width=\columnwidth]{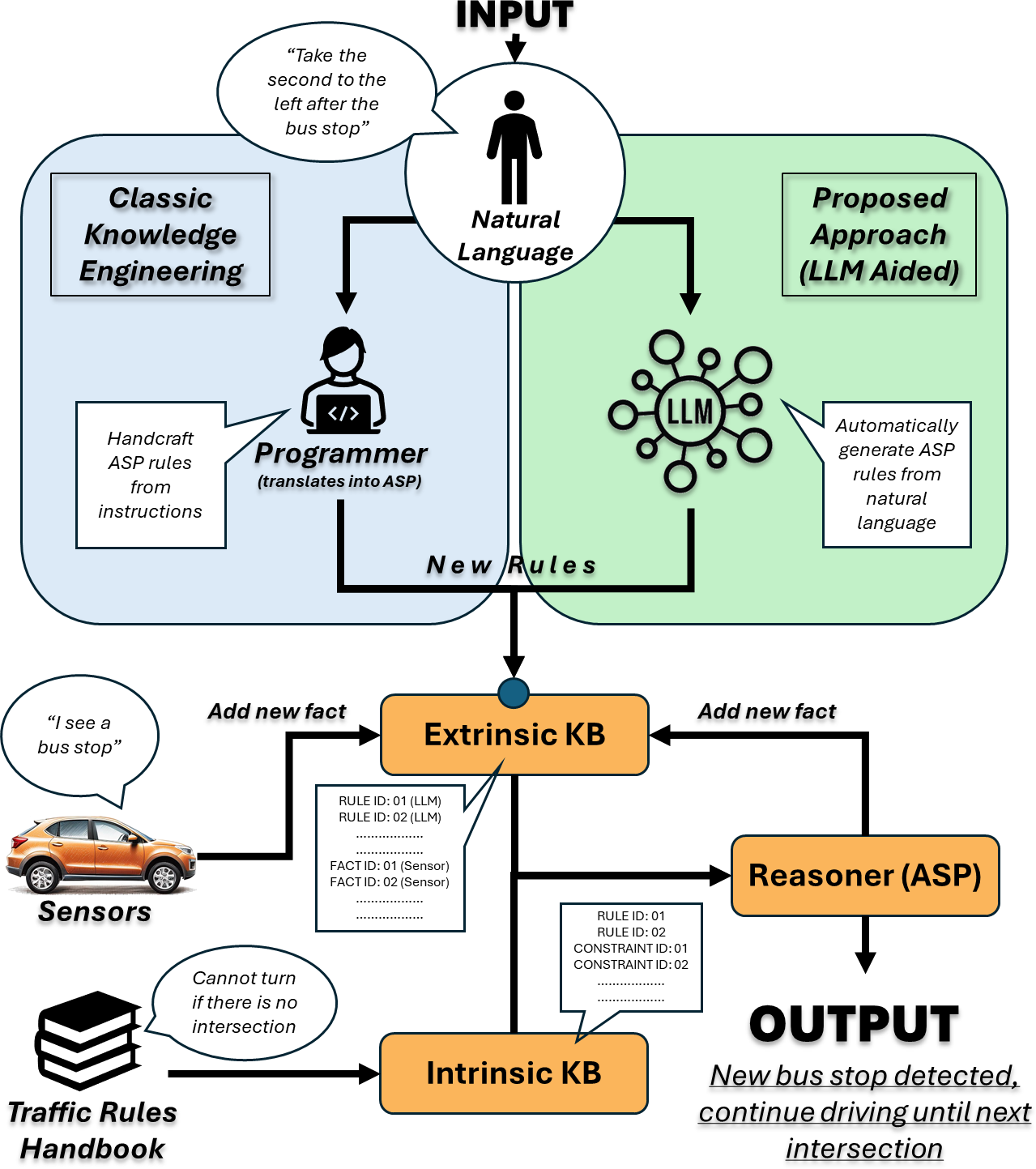}
% \caption{Overview of the HUMANAV pipeline. The focus of this work is highlighted with dashed lines, addressing the knowledge engineering bottleneck by enabling the transition from informal driving instructions to structured ASP rules for reasoning-driven navigation.}
\caption{Overview of the proposed pipeline. The figure shows two flows for generating ASP rules: a classical path via a programmer and an LLM-assisted path. This work focuses on automating this translation to overcome the knowledge engineering bottleneck.}
\label{fig:intro}
\end{figure}
\setlength{\belowcaptionskip}{-10pt}
\begin{figure*}[t]
\centering
\includegraphics[width=\textwidth]{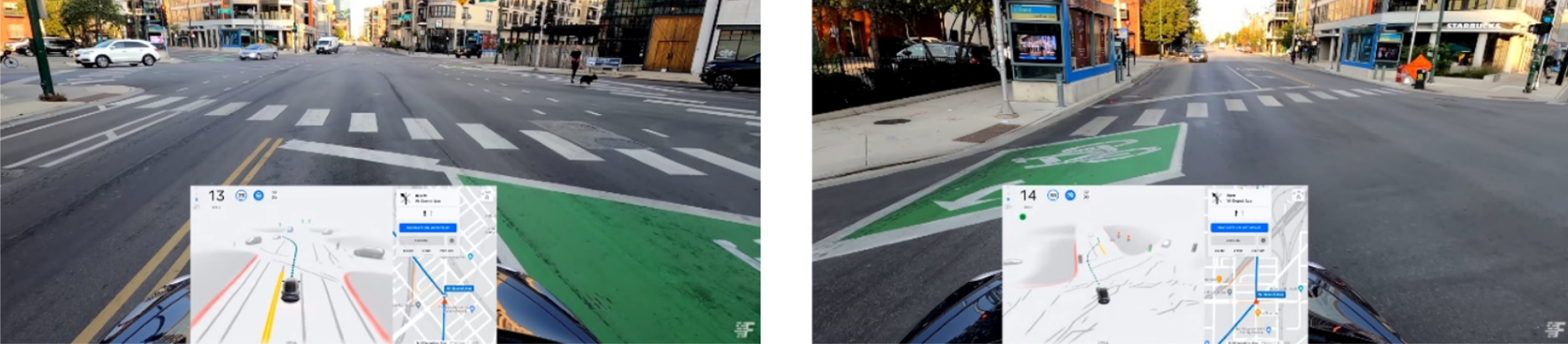}
%\caption{Scene interpretation error. The vehicle control system is unable to correctly recognize the intersection indicated on the map and, as a result, enters the road to the left in the wrong direction. Images extracted from \cite{FSD2021}.}
\caption{Scene interpretation error. The vehicle control system fails to correctly recognize the intersection shown on the map and, as a result, enters the left-hand road in the wrong direction. Images adapted from~\cite{FSD2021}.}
\label{fig:one}
\end{figure*}
The transition to full automation demands robustness in all circumstances, including narrow streets that are difficult to represent using conventional mapping techniques. In 2019, an estimated $22\,700$ people died in traffic accidents in Europe \cite{EC2020}, with 19.4\% of fatalities occurring in environments similar to intersections and 38\% in urban areas \cite{EC2018}. Similar trends have been reported by the Spanish Traffic Authority \cite{DGT2020}, highlighting the significant potential for an autonomous driving system capable of recognizing complex urban environments to enhance road safety and reduce casualties. However, despite the progress made, the current state of technology is still far from achieving full automation as defined by the Society of Automotive Engineers (SAE) LEVEL 5~\cite{SAE2021}, where all driving tasks are automated and can be carried out reliably by the vehicle with no human intervention. Current approaches remain fundamentally dependent on predefined cartographic maps and metric-based localization. 
This work proposes an innovative navigation system for autonomous vehicles that leverages detected environmental features rather than relying on predefined cartographic maps. 
The goal is to enable semantic navigation, moving beyond a purely metric-based approach.
The system supports navigation that more closely resembles human experience, without being strictly dependent on static cartographic material, while adapting to unexpected detours and changes.
Among its advantages, this system include its adaptability to frequent modifications in the road network.
For example, if an intersection is replaced by a roundabout, the system would still be able to process a high-level instruction such as \emph{turn right}, adjusting its understanding dynamically.
To achieve this, the system is designed to mimic human behavior during vehicle operation. By emulating how humans navigate, it introduces a clear separation between vehicle control and environmental perception at a more abstract, semantic level. 
This distinction allows the vehicle’s actions to be guided by reasoning rather than predefined metric-based navigation, enabling adaptation to unexpected conditions and improving robustness in complex road settings.
A key aspect of this approach is the use of \ac{ASP}, a declarative logic-based framework for reasoning-driven navigation. Falling under the broader umbrella of \ac{KR}, ASP enables vehicles to flexibly interpret real-world scenarios and adjust navigation plans on the fly. 
\ac{ASP} specifically enables non-monotonic reasoning, which means it can update conclusions dynamically as new information becomes available. This adaptability is crucial for autonomous systems, allowing them to respond intelligently to unexpected road changes without relying pre-defined maps. 
However, structuring and automating navigation knowledge remains a challenge, commonly referred to as the knowledge engineering bottleneck. \Cref{fig:intro} illustrates this problem, showing the transition from manual rule definition to an automated approach using LLMs to generate ASP rules from natural language instructions.
In this work, we focus on verifying the ability of \acp{LLM} to generate \ac{ASP} code for autonomous navigation. Specifically, we investigate whether informal driving instructions can be translated into structured ASP rules and constraints. By leveraging AI-driven translation, we assess the feasibility of semantic  navigation without relying on predefined cartographic maps, aiming to improve adaptability in dynamic urban environments.

\section{Related Work}
Today, \acp{GNSS} remain the most widely used tools for vehicle positioning, providing drivers with routing assistance through cartographic maps. 
However, they often struggle with signal interference, outdated map data and imprecise localization, leading to navigation errors. 
To address these limitations, research has focused on analyzing camera images and 3D point clouds, which are used to create virtual representations of the vehicle’s surroundings, typically combined with cartographic data to enhance localization.
Yet, in highly dynamic urban environments, unexpected obstacles and changes in road layouts can make reliance on maps and predefined virtual representations insufficient for effective autonomous navigation.
\Cref{fig:one} shows an example in which Tesla's AutoPilot system misinterpreted its position within an intersection, entering the wrong way on the left street. 
Another common case: today, if an intersection is replaced by a roundabout, an automatic system using only cartography struggles to handle this situation.
Similarly, the lack of updated maps would render any map-based autonomous driving system almost useless. 
In such cases, a human driver would typically try to follow the direction of travel semantically, choosing the appropriate exit from the roundabout or following the high-level directions provided by a copilot or external user, such as \emph{``follow the road and take the third exit at the roundabout''}. 
\setlength{\belowcaptionskip}{-14pt} % Adjust as needed
\begin{figure}[t]
  \centering
  % see humanav.mlx on matlab to edit these
  \subfloat[\label{fig:scenario_1a}]{\includegraphics[width=.48\columnwidth]{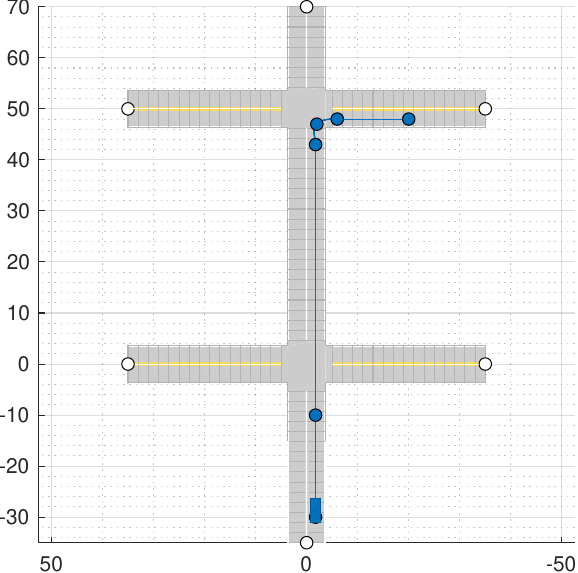}}
  \hfill
  \subfloat[\label{fig:scenario_1b}]{\includegraphics[width=.48\columnwidth]{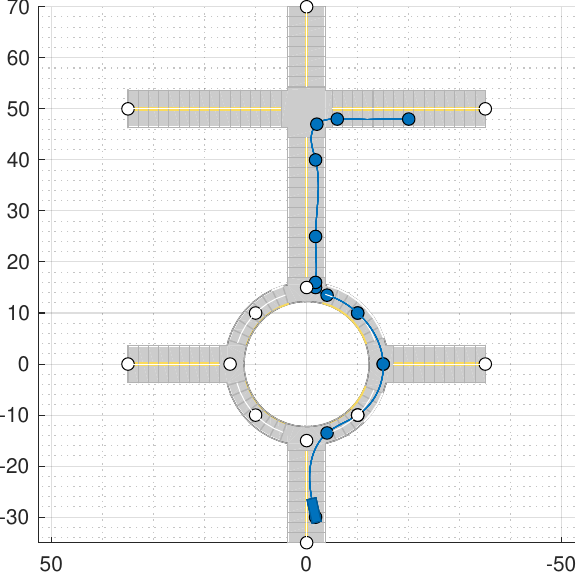}}\\
  \caption{Possible scenarios to reach a gas station, given the instructions.}
  \label{fig:scenario_1}
\end{figure}
\setlength{\belowcaptionskip}{-6pt} % Adjust as needed
This is the type of behavior our approach aims to emulate.
A similar approach to our proposal can be found in \cite{Roh2019Conditional}, where the authors highlight the importance of autonomous vehicles effectively interacting with human users. Their system follows natural-language directions and bypasses challenges related to outdated maps by leveraging images and linguistic input. However, despite this human-centered concept, their approach remains an end-to-end solution without emphasis on explainability.
With the growing use of \acp{LLM}, several recent studies have explored their integration into autonomous driving.
In \cite{ma2024lampilot}, the authors fine-tuned a LLaMA-7B model using the LoRA technique \cite{hu2021lora} to generate highway driving tasks. Their approach, trained on 10k manually annotated driving scenarios, demonstrated higher accuracy than GPT-3.5, validating the effectiveness of their fine-tuning.
Similarly, \cite{chen2024drivingwithllms} created 10k simulated driving scenarios with 160k question-answer pairs for control commands, integrating \acp{LLM} into autonomous driving simulations and evaluating responses in an open-loop system.
The work in \cite{10588851} presents one of the closest ideas to our proposal, where \acp{LLM} generate Python-based driving policies that integrate with traditional planners or controllers. While this framework enables vehicles to follow user instructions in natural language, it relies heavily on LLM reasoning capabilities. Despite the promise of these models and their demonstrated ``sparks of intelligence'' \cite{bubeck2023sparksartificialgeneralintelligence}, their responses still lack rigorous provability, raising reliability concerns for real-world deployment under strictly rigorous terms. 

Building on prior advancements, we introduce a semantic navigation system that emphasizes explainability and adaptability in dynamic urban environments. Unlike existing work that primarily leverages \acp{LLM} to generate control commands or high-level driving policies, our approach explores their potential to produce declarative, logic-based navigation rules using \ac{ASP}. As a well-established \ac{KR} framework, ASP enables a shift from heuristic-driven planning to formal reasoning in autonomous navigation.
In contrast to purely data-driven methods, our ASP-based system ensures that navigation rules are not only derived from human-like reasoning but are also logically verifiable. 
%This distinction enhances the system’s ability to interpret high-level semantic instructions while maintaining the formal constraints necessary for safe operation in complex urban scenarios.
This enhances the system’s ability to interpret high-level instructions while ensuring safe operation in complex urban scenarios.

\section{Proposed Method}
We address the problem of providing navigation instructions to a self-driving system in natural language form. The challenge lies not only in understanding informal driving instructions that are intrinsically vague but also in formalizing them for effective use by the system.
These instructions allow for human-like navigation, such as \textit{``turn left at the bus stop''} instead of \textit{``in 200 meters, turn left''}, reflecting how people naturally think, not in terms of precise distances, but in relation to recognizable waypoints. 
This approach also enables the system to adapt to situations when an intersection is replaced by a roundabout.

% AAAAAAA FUTURE: Although in this work we rely on a set of rules manually created using commonsense reasoning, we can foresee that such driving rules may eventually be inferred automatically by large language models \acp{LLM}, starting from traffic regulations. 

\subsection{Semantic Navigation using ASP}
We focus on translating informal navigation instructions into a structured representation using \ac{ASP}, defined as a set of rules and constraints combined with a complementary program that encodes road-driving rules. 
This transformation allows the system to perform cognitively inspired navigation, interpreting vague natural language instructions in a way that respects traffic regulations and safety constraints.
We illustrate our approach with the following scenario. 
Suppose road construction has significantly changed the network, leaving our navigation system without an accurate map. We need to find a gas station and ask a passerby for directions. The passerby advises: \textit{``turn right after the first intersection''}.
Confident in our understanding, we proceed, mentally picturing the situation in \Cref{fig:scenario_1a}. 
After driving for a while, we encounter a roundabout that was not mentioned. Rather than being discouraged, we assume the intersection was replaced and adjust our mental model accordingly (\Cref{fig:scenario_1b}). We enter the roundabout and select the exit that best aligns with the intended direction of \textit{``straight ahead''}.
\begin{figure}[t!]
\begin{lstlisting}[numbers=left,xleftmargin=10pt,framexleftmargin=10pt, stepnumber=1, language=Prolog, basicstyle=\footnotesize\ttfamily,keepspaces=true, morekeywords={T, T1, T2, Time, ID, ID1, ID2, Time1, Time2, maxint}, caption={DLV constraints generated for Experiment ID1}, captionpos=t, label={lst:chatgpt_answer_1},xleftmargin=10pt,framexleftmargin=0em, numbersep=2pt,aboveskip=0pt,belowskip=-15pt]
% The car must turn left exactly once.
:-#count{T : cross_left(T)}=0.
:-#count{T : cross_left(T)}>1.

% The car must cross straight exactly once.
:-#count{T : cross_straight(T)}=0.
:-#count{T : cross_straight(T)}>1.

% The car must turn right exactly once.
:-#count{T : cross_right(T)}=0.
:-#count{T : cross_right(T)}>1.

% The left turn must happen first.
:-cross_left(T1), cross_straight(T2), T1>=T2.
:-cross_left(T1), cross_right(T3), T1>=T3.

% The straight crossing must happen before 
% the right turn.
:-cross_straight(T2), cross_right(T3),T2>=T3.
\end{lstlisting}
\end{figure}
After continuing for a short distance, we reach a second intersection where we turn right and successfully reach the gas station, relieved to have avoided running out of fuel.
While this scenario is simple, it demonstrates the flexibility needed in autonomous navigation, a capability that existing navigation systems lack, making them far from achieving SAE Level 5 automation.
Before addressing how our approach navigates from the initial position to the gas station, we must first recognize that we are essentially facing a planning problem. 
In a stable, unchanging road environment where maps are consistently updated, this problem is trivial and already solvable with vehicle navigation systems. However, in our scenario, maps are unreliable, requiring an alternative approach.
A straightforward solution might involve combining a scene understanding system with a traditional imperative program, designed with extensive knowledge to recognize various driving situations. This could involve defining an abstract control flow, adaptable to specific navigation needs.
Instead, we propose solving the navigation task using an \ac{ASP} model, describing the problem through logical rules that define the result rather than the process.
Similar to the imperative program approach, we still feed our system with semantic information extracted from scene understanding systems. 
However, in this case, the information is treated as facts that \ac{ASP} processes to generate all possible \textit{``answer-sets''}, that is, all solutions that satisfy our requirements, ultimately providing high-level instructions to the vehicle’s low-level controller.
As the vehicle moves, new environmental facts are collected, potentially invalidating previous plans. This dynamic behavior is a core advantage of ASP’s \textbf{non-monotonic reasoning}, which enables the system to dynamically adjust to incomplete or changing information. 

\subsection{Knowledge Engineering with LLMs}
To evaluate our approach in practice, we designed a structured navigation task described in the next subsection. 
Building on the representation introduced earlier, we explore the potential of \acp{LLM} to automate the generation of navigation-specific ASP rules. Specifically, we investigate whether informal driving instructions can be systematically translated into a consistent set of rules and constraints compatible with the DLV system, a widely used \ac{ASP} solver.
To support reasoning and navigation, we define a structured \ac{KB} composed of logical rules and environmental facts. 
This \ac{KB} captures both general traffic behavior and situation-specific navigation goals.
Within our system, the \ac{KB} is divided into two parts. First, an \textit{intrinsic} \ac{KB} is manually designed, containing general traffic rules and commonsense driving behaviors. These rules encode basic legal principles derived from traffic legislation, such as crossing intersections, navigating roundabouts, and obeying right-of-way conventions. 
Importantly, this intrinsic knowledge is fixed and independent of any specific navigation instruction, ensuring that the vehicle always operates in accordance with standard road regulations.
Conversely, an \textit{extrinsic} \ac{KB} must be generated dynamically, depending both on the specific navigation instruction received and the real-time perception of the environment. For example, from the instruction \textit{``turn right after the first
intersection''}, the system must produce formal constraints representing the intended sequence of maneuvers. 
At the same time, it must integrate sensor detections, including intersections, roundabouts, and straight roads, updated continuously as the vehicle progresses.
Both the goal-specific constraints and the sensor detections are combined within the extrinsic \ac{KB} to guide the reasoning process.
We hypothesize that \acp{LLM} can assist in automatically generating part of the extrinsic \ac{KB} by interpreting informal human instructions and translating them into formal DLV constraints compatible with the intrinsic traffic rules.

\section{Experimental Validation}
\label{sec:Experimental-Validation}
This section describes the experiments we designed to validate the feasibility and robustness of the proposed LLM-assisted knowledge engineering approach.
Our objective is to evaluate whether different \acp{LLM} can reliably generate DLV code from informal driving instructions, capturing the intended sequences of driving maneuvers.+

We tested multiple \acp{LLM} on two different tasks, progressively increasing complexity to highlight the capability of the DLV framework to model different navigation scenarios.
Given the navigation instructions in \Cref{tab:experimental-set}, we provided a detailed prompt to all listed LLMs evaluating whether models were able to generate constraints that correctly formalize the specified driving instructions.
The prompt included our handbook of driving rules, the description of the navigation task in natural language, DLV-specific syntax guidelines such as the handling of time variables, the absence of a direct $\neq$ operator and the use of aggregate constraints for counting events.
Our basic yet effective driving rules include information about the road graph, such as how road segments can be connected both by roundabouts and intersections.

\begin{figure}[t]
\centering
\includegraphics[width=\columnwidth]{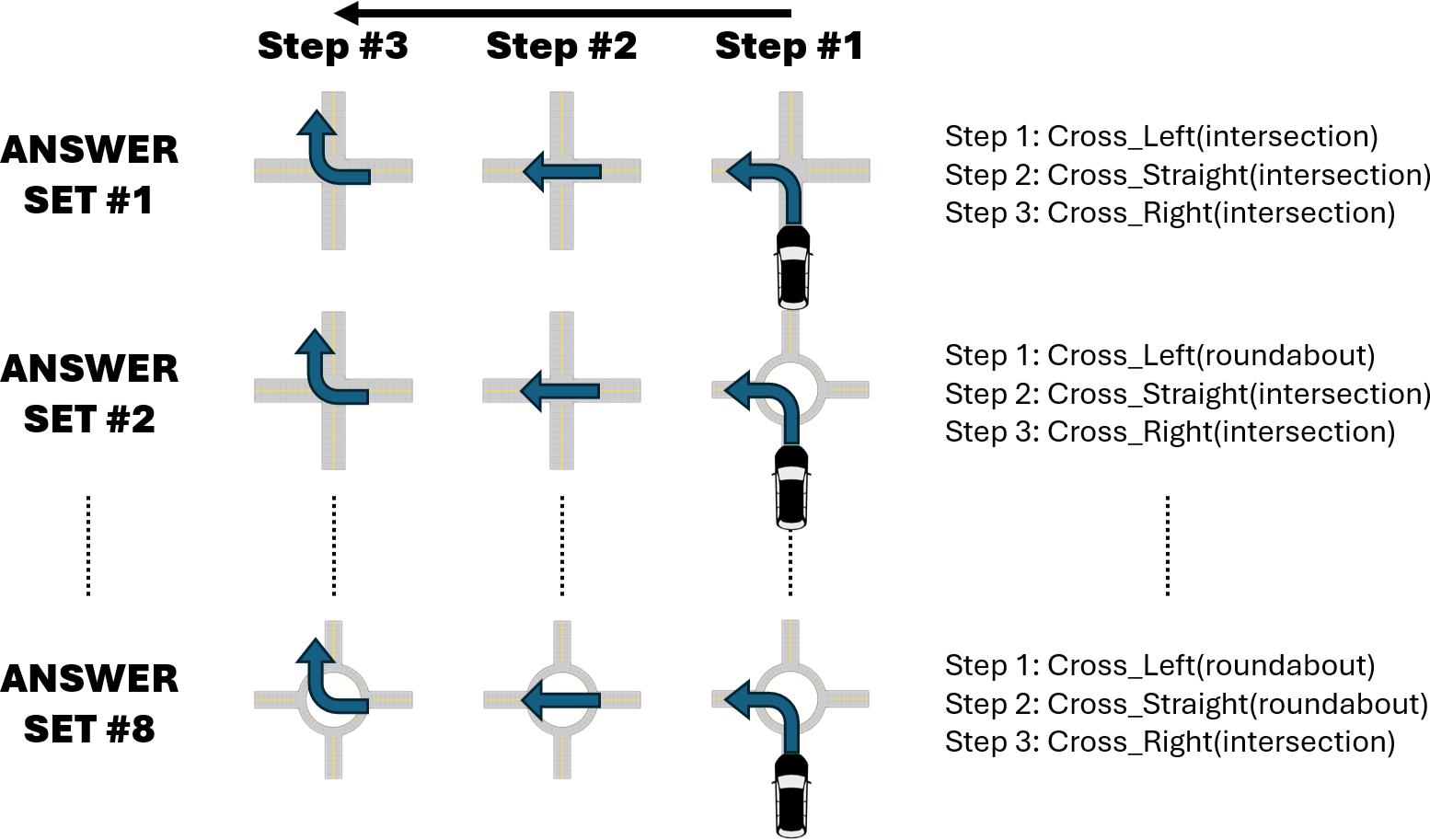}
\caption{Possible scenarios derived from \textit{``turn left at the next junction, then go straight, and finally turn right''}.}
\label{fig:8cases}
\end{figure}

\subsection{Experimental Setup}
\label{sec:Experimental-Setup}
We evaluated the set of experiments set using ChatGPT-4o, Grok-3, Groq Compound Beta Mini, Gemini-2.0-Flash, Meta Llama 4, Qwen2.5-Max, and DeepSeek for experiments ID0 to ID2, and ChatGPT-4.5 for experiment ID3.
Without loss of generality, we assume a road network made of four-exit roundabout and standard aligned intersections. More complex or asymmetric geometries could be accommodated in future work by extending the intrinsic \ac{KB}.
We define \textit{success} using two measurable criteria: syntactic validity and semantic correctness. 
Syntactic validity requires that the LLM-generated code be correctly parsed by DLV without errors, while semantic correctness is satisfied when the integrated ASP model produces answer sets consistent with the intended sequence of driving 
maneuvers.
From a technical perspective, the execution flow proceeds as follows. Given the handbook of driving rules, which represents the intrinsic knowledge base, we simulate all possible sensors readings to construct the extrinsic knowledge base for each experiment. For instance, given experiment ID0, we simulate the finite set of junctions including intersections and roundabout entities necessary to achieve the instructions.
Experiments consist of adding LLM-generated constraints to the baseline rule set to limit the combinatorial explosion of the aforementioned rules. 
This execution flow allow us to build a self-contained testbed that does not rely on external inputs or simulation environments.
For example, given the instructions in ID1: \textit{``turn left at the next junction, then go straight, and finally turn right''}, the constraints generated by the LLM (shown in \Cref{lst:chatgpt_answer_1}) generate the expected 8 answer set illustrated in \Cref{fig:8cases}. Without these constraints, the handbook of driving rules alone would create 463 answer sets. It must be noted that DLV does not compute all answer sets exhaustively, but rather uses optimization techniques to avoid redundant computation. In this case, the execution time was 0.023 seconds on an Intel core i9-14900k machine. 
Finally, we report in the tables also the number of lines of code generated by each LLM. 
We have found this useful as a measure of actual understanding of the models, since LLMs often produce redundant or unused rules. 
Results are presented in \Cref{tab:experimental-results00,tab:experimental-results01,tab:experimental-results02}.

\begin{table}[]
\centering
\caption{List of navigation instructions used in the experimental validation, ordered by increasing complexity.}
\label{tab:experimental-set}
\begin{tabular}{|c|c|c|}
\hline
\textbf{ID} & \textbf{Task} & \textbf{Instruction} \\ \hline
ID0 & Add constraints & \begin{tabular}[c]{@{}c@{}}Go straight at the first junction, \\ then turn right.\end{tabular} \\ \hline
ID1 & Add constraints & \begin{tabular}[c]{@{}c@{}}Turn left at the next junction, \\ then go straight, then turn right.\end{tabular} \\ \hline
ID2 & Add constraints & Turn right after the first junction \\ \hline
ID3 & \begin{tabular}[c]{@{}c@{}}Generate Rules \\ and constraints\end{tabular} & Go straight at the next intersection. \\ \hline
\end{tabular}
\end{table}
\begin{table}[]
\centering
\caption{Experiment ID0, as validation criteria in \Cref{sec:Experimental-Setup}.}
\label{tab:experimental-results00}
\begin{tabular}{|l|c|c|c|}
\hline
\multicolumn{1}{|c|}{\textbf{LLM Model}} & \textbf{Syntax} & \textbf{Semantic} & \multicolumn{1}{l|}{\textbf{Lines}} \\ \hline
ChatGPT4-o & \ding{51} & \ding{51} & 7 \\ \hline
Grok 3 & \ding{51} & \ding{51} & 8 \\ \hline
Gemini-2.0-Flash & \ding{51} & \ding{51} & 6 \\ \hline
Meta Llama4 & \ding{51} & \ding{51} & 8 \\ \hline
Qwen2.5-Max & \ding{55} & \ding{55} & - \\ \hline
Groq Compound Beta Mini & \ding{51} & \ding{55} & - \\ \hline
DeepSeek & \ding{55} & \ding{55} & - \\ \hline
\end{tabular}
\end{table}
\begin{table}[]
\centering
\caption{Experiment ID1, as validation criteria in \Cref{sec:Experimental-Setup}.}
\label{tab:experimental-results01}
\begin{tabular}{|l|c|c|c|}
\hline
\multicolumn{1}{|c|}{\textbf{LLM Model}} & \textbf{Syntax} & \textbf{Semantic} & \multicolumn{1}{l|}{\textbf{Lines}} \\ \hline
ChatGPT4-o & \ding{51} & \ding{51} & 12 \\ \hline
Grok 3 & \ding{51} & \ding{51} & 8 \\ \hline
Gemini-2.0-Flash & \ding{51} & \ding{51} & 11 \\ \hline
Meta Llama4 & \ding{51} & \ding{51} & 8 \\ \hline
Qwen2.5-Max & \ding{51} & \ding{51} & 10 \\ \hline
Groq Compound Beta Mini & \ding{55} & \ding{55} & - \\ \hline
DeepSeek & \ding{55} & \ding{55} & - \\ \hline
\end{tabular}
\end{table}
\begin{table}[]
\centering
\caption{Experiment ID2, as validation criteria in \Cref{sec:Experimental-Setup}.}
\label{tab:experimental-results02}
\begin{tabular}{|l|c|c|c|}
\hline
\multicolumn{1}{|c|}{\textbf{LLM Model}} & \textbf{Syntax} & \textbf{Semantic} & \multicolumn{1}{l|}{\textbf{Lines}} \\ \hline
ChatGPT4-o & \ding{51} & \ding{51} & 6 \\ \hline
Grok 3 & \ding{51} & \ding{51} & 8 \\ \hline
Gemini-2.0-Flash & \ding{55} & \ding{55} & - \\ \hline
Meta Llama4 & \ding{51} & \ding{51} & 8 \\ \hline
Qwen2.5-Max & \ding{51} & \ding{51} & 9 \\ \hline
Groq Compound Beta Mini & \ding{55} & \ding{55} & - \\ \hline
DeepSeek & \ding{55} & \ding{55} & - \\ \hline
\end{tabular}
\end{table}
\subsection{Results on Structured Navigation Instructions}
\label{sec:Results-on-Structured-Navigation-Instructions}
\begin{figure}[t]
\centering
\includegraphics[width=\columnwidth]{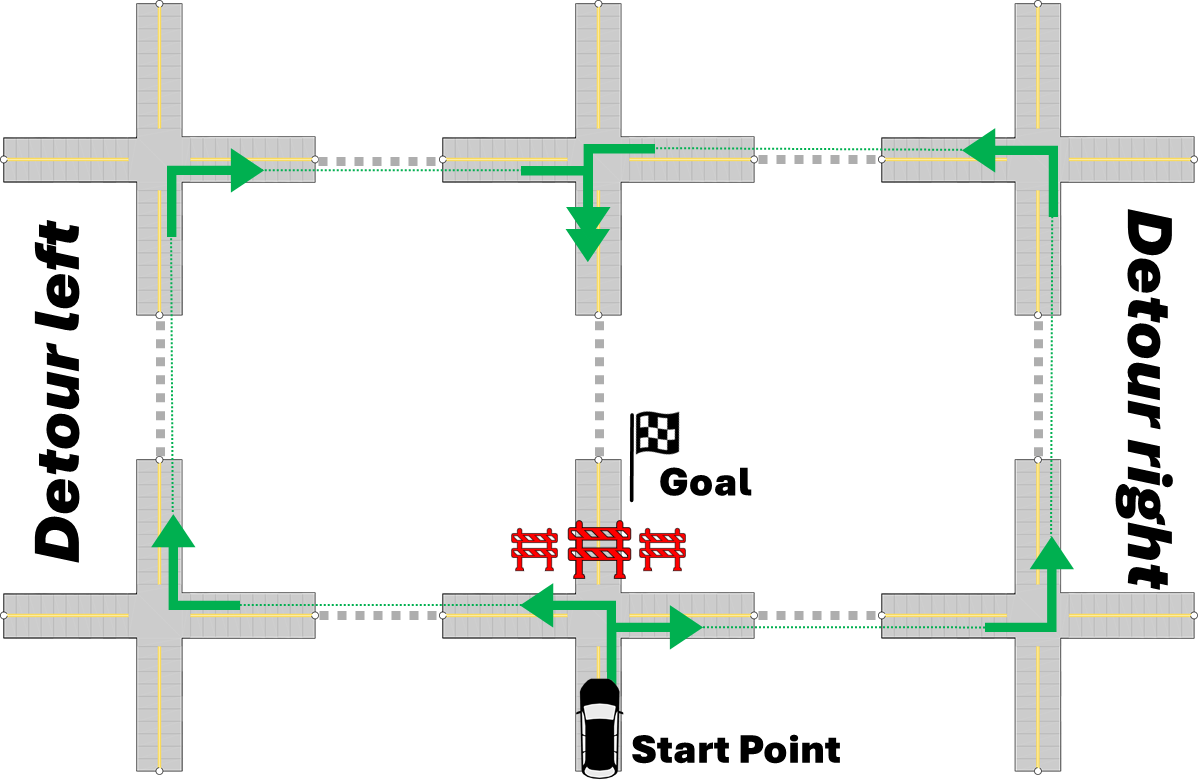}
\caption{Experiment ID3. The black car indicates the starting position. Green arrows represent the two navigation scenarios modeled by the DLV-generated rules.}
\label{fig:ID3}
\end{figure}

The first set of experiments, between ID0 and ID2, were designed to test the feasibility of automatically generating navigation constraints.
When integrated with the intrinsic knowledge base, model execution confirmed that the resulting answer sets satisfied the intended sequence of maneuvers without conflicts or inconsistencies.
Our results show that, in the absence of syntax errors, most LLMs produced comparable outputs. The only semantic failure occurred in the ID0-Groq case, where the model generated constraints that led to inconsistency (i.e., no valid answer set). 
From a technical perspective, this suggests that current LLMs can grasp DLV syntax (despite its limited exposure compared to more common programming languages like Python or C), and can translate simple instructions into valid DLV logic programs.
In experiment ID3, we extended the task by asking the model to generate not only constraints but also logic rules to handle a detour.
The prompt included the traffic rule handbook, the ID3 instruction, and a sentence specifying to prepare a new program including new rules to handle the detour. 
For this test, we used the most advanced ChatGPT4.5 model from OpenAI. 
The model produced 47 rules and constraints that correctly modeled the intended scenario (see \Cref{fig:ID3}).
It must be said that, if previous experiment supports the kind of ``sparks of intelligence'' described in~\cite{bubeck2023sparksartificialgeneralintelligence}, we also tested whether a valid DLV model could be produced without providing the handbook of traffic rules. In this case, several issues emerged, including invalid syntax (due to mixing different ASP dialects) and contradictory rules.
Nevertheless, these experiments demonstrate that current LLMs can effectively support the formalism required by the DLV symbolic reasoner, paving the way for more advanced \ac{RAG} systems~\cite{HUSSIEN2025125914} that leverage the explainability inherent in ASP-based reasoning.

%ChatGPT experiment "ITSC2015-Experiment ID03"
%"I want the goal is to go straight at the first intersection and then stop. But it can be the case that there is a detour so we cannot go straight at the first intersection. In this case we have to go around the detour. We assume the road network to be connected with multiple intersections of 4 ways."

% \subsection{Adaptability to Environmental Changes}
% - % let keep this more fancy things for the journal
% - Describe how reasoning dynamically handles changes (missed exits, blocked roads) without modifying original instruction.
% - Show that ASP non-monotonicity recomputes valid plans as new facts arrive.
\setlength{\belowcaptionskip}{-10pt}
\section{Discussion and Future Work}
This work presented an integration of \acfp{LLM} with a \acf{KR} system that simultaneously aim to: (1) decouple navigation from existing cartography, adding a layer of reliability in dynamic environments; (2) facilitate a human-like driving experience, where driving instructions can be provided in natural language form; and (3) intrinsically provide a layer of explainability through the traceable reasoning contained in the generated answer sets.
While our current implementation relies on manually defined driving rules based on commonsense reasoning, we envision that such rules could eventually be inferred automatically from formal traffic regulations using \acp{LLM}. We also anticipate that knowledge-graph-based mechanisms, capturing past driving experiences, could enhance the quality and contextual relevance of LLM-generated constraints. Additionally, future work will address the integration of a closed-loop simulation environment enabling validation in dynamic conditions. Finally, we plan to extend our system to handle more complex intersection geometries, further broadening its applicability to real-world autonomous navigation tasks.

\newcommand{\myfontsize}{\scriptsize}    % Font size (default: \tiny or \scriptsize or \footnotesize)
\newcommand{\myleftpadding}{0pt}   % Left padding (default: 2pt)
\newcommand{\myrightpadding}{0pt}  % Right padding (default: 2pt)
\newcommand{\mytitlefont}{\bfseries} % Title font (default: \bfseries for bold)

\section*{Acknowledgments}
\noindent This research has been partially funded by the María Zambrano Grants and the PIUAH23/IA-039 project granted by Universidad de Alcalá. ALB would like to thank Domenico Sorrenti and Matteo Palmonari for their help and support.


\begin{thebibliography}{1}

\bibitem{WaymoSupport}
Waymo Support, ``Waymo Help Center,'' \url{https://support.google.com/waymo/answer/9775645}.% [Accessed: 01-Oct-2024].

\bibitem{SAE2021} 
SAE, ``SAE Levels of Driving Automation,'' 2021. \url{https://www.sae.org/blog/sae-j3016-update}.% [Accessed: 01-Oct-2024].

\bibitem{EC2020}
European Commission, ``Road safety key figures 2020,''.  \url{https://ec.europa.eu/transport/road_safety/sites/roadsafety/files/pdf/scoreboard_2020.pdf}.%[Accessed: 01-Oct-2024].

\bibitem{EC2018}
European Commission, ``European union annual accident report 2018,'' \url{https://ec.europa.eu/transport/road_safety/specialist/observatory/statistics/annual_accident_report_archive_en}.% [Accessed: 01-Oct-2024].

\bibitem{DGT2020}
Dirección General de Tráfico (DGT), ``Avance de las principales cifras,'' 2020. [Online]. Available: \url{https://www.dgt.es/export/sites/web-DGT/.galleries/downloads/dgt-en-cifras/24h/Las-principales-cifras-2020_v6.pdf}.% [Accessed: 01-Oct-2024].

\bibitem{Habermann2016}
D. Habermann, C. E. Vido, F. S. Osório, and F. Ramos, ``Road junction detection from 3D point clouds,'' in IJCNN2016. %*International Joint Conference on Neural Networks (IJCNN)*, Vancouver, BC, Canada, 2016.

\bibitem{Augusto2017}
A. L. Ballardini, D. Cattaneo, S. Fontana, and D. Sorrenti, ``An online probabilistic road intersection detector,'' in ICRA2017. %*2017 IEEE International Conference on Robotics and Automation (ICRA)*, 2017.

\bibitem{Augusto2021}
A. L. Ballardini, Á. Hernández, S. Carrasco, J. Lorenzo, I. Parra, N. Hernández, I. García, and M.A. Sotelo, ``Urban intersection classification: A comparative analysis,''. % *Sensors*, vol. 21, p. 6269, 2021.

\bibitem{FSD2021}
``FSD Beta v10.0 First Drive \& Impressions 2021.24.15 FSD Beta 10,''  \url{https://youtube.com/clip/UgkxS2jDErt524J7FPvSjdXU6qYVELVYyB_z}.% [Accessed: 01 October-2024].

\bibitem{Roh2019Conditional}
J. Roh, C. Paxton, A. Pronobis, A. Farhadi, and D. Fox, ``Conditional driving from natural language instruction,'' in Proceedings of the Conference on Robot Learning, 2019.

\bibitem{ma2024lampilot}
Y. Ma, C. Cui, X. Cao, W. Ye, P. Liu, J. Lu, A. Abdelraouf, R. Gupta, K. Han, A. Bera, J. M. Rehg, and Z. Wang, ``LaMPilot: An open benchmark dataset for autonomous driving with language model programs,'' in CVPR2024. %*Proceedings of the IEEE/CVF Conference on Computer Vision and Pattern Recognition (CVPR)*, 2024.

\bibitem{chen2024drivingwithllms}
L. Chen, O. Sinavski, J. Hünermann, A. Karnsund, A. J. Willmott, D. Birch, D. Maund, and J. Shotton, ``Driving with LLMs: Fusing object-level vector modality for explainable autonomous driving,'', ICRA2024. % in *2024 IEEE International Conference on Robotics and Automation (ICRA)*, 2024.

\bibitem{10588851}
Y. Wang, Z. Huang, Q. Liu, Y. Zheng, J. Hong, J. Chen, L. Xiong, B. Gao, and H. Chen, ``Drive as veteran: Fine-tuning of an onboard large language model for highway autonomous driving,'' in IV2024. %*, pp. 502-508, 2024. doi: 10.1109/IV55156.2024.10588851.

\bibitem{hu2021lora}
E. J. Hu, Y. Shen, P. Wallis, Z. Allen-Zhu, Y. Li, S. Wang, L. Wang, and W. Chen, ``LoRA: Low-rank adaptation of large language models,'' %*arXiv preprint arXiv:2106.09685*, 2021.

\bibitem{bubeck2023sparksartificialgeneralintelligence}
S. Bubeck, V. Chandrasekaran, R. Eldan, J. Gehrke, E. Horvitz, E. Kamar, P. Lee, Y. T. Lee, Y. Li, S. Lundberg, H. Nori, H. Palangi, M. T. Ribeiro, and Y. Zhang, ``Sparks of artificial general intelligence: Early experiments with GPT-4,'' %*arXiv preprint arXiv:2303.12712*, 2023. [Online]. Available: \url{https://arxiv.org/abs/2303.12712}.

\bibitem{FIKES1971189}
R. E. Fikes and N. J. Nilsson, ``STRIPS: A new approach to the application of theorem proving to problem solving,'' \textit{Artificial Intelligence}.%, vol. 2, no. 3, pp. 189-208, 1971. %doi: \href{https://doi.org/10.1016/0004-3702(71)90010-5}{10.1016/0004-3702(71)90010-5}.

\bibitem{DLV}
N. Leone, G. Pfeifer, W. Faber, T. Eiter, G. Gottlob, S. Perri, and F. Scarcello, ``The DLV system for knowledge representation and reasoning,'' \textit{ACM Transactions on Computational Logic}.%, vol. 7, no. 3, pp. 499–562, Jul. 2006. %doi: \href{https://doi.org/10.1145/1149114.1149117}{10.1145/1149114.1149117}.

\bibitem{HUSSIEN2025125914}Hussien, M., Melo, A., Ballardini, A., Maldonado, C., Izquierdo, R. \& Sotelo, M. RAG-based explainable prediction of road users behaviors for automated driving using knowledge graphs and large language models. %{\em Expert Systems With Applications}. %\textbf{265} pp. 125914 (2025). %https://www.sciencedirect.com/science/article/pii/S0957417424027817


\end{thebibliography}
\end{document}